\documentclass{article}
\usepackage{iclr2027_conference,times}

\usepackage{amsmath,amsfonts,bm}

\def\eqref#1{equation~\ref{#1}}

\def\1{\bm{1}}

\DeclareMathAlphabet{\mathsfit}{\encodingdefault}{\sfdefault}{m}{sl}
\SetMathAlphabet{\mathsfit}{bold}{\encodingdefault}{\sfdefault}{bx}{n}

\usepackage{amsmath,amssymb,booktabs,graphicx,array}
\usepackage{flafter}
\usepackage[section]{placeins}
\usepackage{hyperref}
\usepackage{url}
\hypersetup{
  pdftitle={First-Layer Localization Functions in Physics-Informed Neural Networks for Long-Domain and High-Order Problems},
  pdfauthor={Anonymous},
  colorlinks=true,
  linkcolor=blue,
  urlcolor=blue,
  citecolor=blue
}
\newcommand{\hd}[1]{\shortstack[r]{#1}}

\newcommand{\best}[1]{\boldmath\textbf{#1}\unboldmath}

\iclrfinalcopy
\author{Lakshay Chawla\\
 Indian Institute of Technology Jodhpur\\
 \texttt{lakshaychawla13@gmail.com}
 \And
 Hardik Jain\\
 Indian Institute of Technology Jodhpur\\
 \texttt{hardik.jain@iitj.ac.in}}

\title{LPINNs: First-Layer Gated Localization for Physics-Informed Neural Networks}

\begin{document}
\maketitle

\begin{abstract}
Physics-informed neural networks (PINNs) use one shared representation over the computational domain, which can become difficult to optimize on long domains and for high-order operators. We study a minimal alternative: multiply the first hidden activation of an otherwise unchanged dense PINN by input-dependent localization functions, giving first-layer units receptive fields without partitioning the domain or adding interface losses. We screen 13 families of localization functions, in up to three parameterizations each, on a nonlinear harmonic oscillator (HO), a heat equation on a long spatial interval, and a manufactured four-dimensional (4D) fourth-order problem, with ten paired seeds throughout. Three configurations give large reductions in solution error at matched budgets: (i) Fixed Gaussian localization functions on the $2\pi$ HO domain cut mean solution RMSE from $4.8369\times10^{-1}$ to $8.83\times10^{-3}$ at 3k epochs. (ii) The inverse-quadratic family with learnable centers and widths cuts it from $2.896\times10^{-1}$ to $3.06\times10^{-2}$ on the $8\pi$ heat domain at 10k epochs. (iii) Fixed bump localization functions cut it from $1.75947\times10^{1}$ to $2.260\times10^{-1}$ on the $4\pi$ 4D domain at 10k epochs. Every paired seed improves in these three comparisons. The screen also shows that the mechanism is not a free win: on HO only 2 of 13 families beat the baseline, and 10 of the remaining 11 are 9 to 23 times worse; on 4D four families are non-finite and five are more than three orders of magnitude worse than the baseline. The inverse-quadratic family is the only one that beats the baseline on all three equations. Overall, these results show that first-layer localization can provide measurable improvements to baseline PINNs on long-domain and high-order problems.
\end{abstract}

\section{Introduction}
A physics-informed neural network replaces a mesh-based solver with a neural approximation trained to reduce a differential-equation residual at collocation points \citep{lagaris1998artificial,raissi2019physics}. Automatic differentiation makes this attractive, but training can stall when the loss is poorly conditioned, the target contains several length scales, or the residual contains high-order derivatives \citep{wang2021understanding,krishnapriyan2021characterizing}. A fully connected network also uses the same first-layer features across the entire coordinate domain. A feature useful near one end of a long interval is therefore active, and updated, at the other end.

One established response is domain decomposition: cPINNs, XPINNs, FBPINNs, and related local methods use several approximators and couple them through partitions, overlaps, or interface conditions \citep{jagtap2020conservative,jagtap2020extended,moseley2023finite,dong2021local}. These methods add useful structure, but also add networks and coupling terms. This work explores whether part of the benefit can be obtained with a single dense PINN.

Our intervention is to multiply the first hidden representation by a bank of input-dependent localization functions. Each first-layer unit then has a receptive field, while all later layers remain dense and global. The residual, optimizer, collocation points, and boundary-condition transform are unchanged, and no subdomain partition or interface term is introduced, so the network still represents one smooth global function.

We report two kinds of evidence. Section~\ref{sec:results} gives three detailed paired-seed comparisons, one per equation, in which localization reduces solution error by one to two orders of magnitude. Section~\ref{sec:family-screen} gives the screen over all 13 families of localization functions at the same settings, which is what makes the detailed comparisons interpretable: most families do not help, several are far worse than the baseline, and four are numerically non-finite on the 4D problem. Taken together, the two views support the following conclusions

\begin{itemize}
\item First-layer localization can substantially improve recovered solution accuracy in the tested long-domain and high-order settings, with the improvement holding in all ten paired seeds of each detailed comparison.
\item The benefits of localization are equation-dependent, with different function families providing the strongest gains on different problems.
\end{itemize}

This is an initial controlled study on one canonical ODE, one canonical PDE, and one manufactured high-order problem, and its comparisons are matched on training budget rather than on computational cost.

\section{Related work}
PINNs were introduced as neural approximations trained through differential-equation residuals and boundary or initial conditions \citep{lagaris1998artificial,raissi2019physics}. A broad survey of the area is given by \citet{karniadakis2021physics}. Their optimization difficulties include imbalanced gradients \citep{wang2021understanding}, stiff training dynamics explained through the neural tangent kernel \citep{wang2022when}, spectral bias on oscillatory targets \citep{rahaman2019spectral,wang2021eigenvector}, and residual minima that do not correspond to accurate solutions \citep{krishnapriyan2021characterizing}.

Local PINN methods address these difficulties by using multiple subnetworks or local basis functions. cPINNs enforce conservation across subdomains, XPINNs generalize the decomposition to space-time, and FBPINNs combine overlapping local networks through a partition of unity \citep{jagtap2020conservative,jagtap2020extended,moseley2023finite}. Local extreme learning machines and compactly supported variational bases follow a related motivation \citep{dong2021local,kharazmi2021hp}, and \citet{heinlein2021combining} review the line of work. Our method keeps one network and introduces locality only through input-dependent modulation of its first hidden layer.

The construction is related to localized units in radial-basis networks \citep{broomhead1988multivariable,park1991universal} and to neuron-wise adaptive activations \citep{jagtap2020locally}, but differs from both: our localization functions depend on the input and are inserted inside a standard multilayer PINN rather than used as the output basis or as input-independent slopes.

\section{First-layer localization}
\label{sec:method}
Let $x\in\mathbb{R}^{d}$ be the physical coordinates and $\widetilde{x}\in[0,1]^d$ their coordinate-wise normalization. A standard multilayer perceptron is
\begin{align}
  h^{(1)}&=\phi(W_0\widetilde{x}+b_0),\\
  h^{(\ell+1)}&=\phi(W_\ell h^{(\ell)}+b_\ell),
\end{align}
followed by a scalar linear output $N_\theta$. Localization replaces only the first hidden representation with
\begin{equation}
  h^{(1)}_{\mathrm{loc}}=h^{(1)}\odot g(\widetilde{x};\theta_g),
  \label{eq:localized}
\end{equation}
where $g$ returns one value per first-layer unit and $\odot$ is elementwise multiplication. We refer to $g$ as a localization function: it multiplicatively modulates the first hidden-layer representation and gives each unit an input-dependent receptive field. The later layers remain ordinary dense layers. For the Gaussian family, the localization function $g_j$ associated with first-layer unit $j$ has centers $\mu_j$ and widths $\sigma_j$ giving
\begin{equation}
  g_j(\widetilde{x})=\exp\left[-\dfrac{1}{2}\sum_{k=1}^{d}
  \left(\dfrac{\widetilde{x}_k-\mu_{kj}}{\sigma_{kj}}\right)^2\right]
  \label{eq:gaussian}
\end{equation}
with centers spread over $[0,1]^d$ and widths set from the center spacing, so that neighboring receptive fields overlap.

\paragraph{What the localization functions change, and what they do not.} The localization field is a smooth function of the input, so the network output remains one globally defined smooth function; there is no partition, no interface penalty, and no continuity constraint to satisfy. What changes is which parameters a given collocation point can move. Differentiating Eq.~\ref{eq:localized} w.r.t. $\widetilde{x}_k$ gives 
\begin{equation}
    \partial_k h^{(1)}_{\mathrm{loc},j}=g_j\,\partial_k h^{(1)}_j+h^{(1)}_j\,\partial_k g_j
\end{equation}
and both terms are small for units whose localization function is negligible at that point. A residual evaluated near one end of the domain, therefore, produces gradients concentrated on the first-layer units whose receptive fields cover that region, rather than on all of them, while the later layers still mix features from every unit. This is a weaker form of locality than domain decomposition and it is correspondingly cheaper: one bank of localization functions, no extra networks, no interface terms, and no per-subdomain hyperparameters.

\paragraph{Why one layer.} We place the localization functions at the first layer for two practical reasons. (i) It is the smallest change that can create receptive fields, and later layers already mix features globally. (ii) Localizing at every layer would also add spatially varying parameters to every derivative order in the residual, including four orders in the 4D problem. The reported experiments therefore characterize a single bank of localization functions; they are not evidence that other placements cannot work.

\paragraph{Localization-function families and parameterizations.} We consider 13 families of localization functions: Gaussian, activated, super-Gaussian, Ricker, boxcar, Laplace, Cauchy, raised-cosine, bump, Gabor, Morlet, inverse-quadratic, and triangular. Each family is tested with fixed centers and widths (F), learnable widths (S), and learnable centers and widths (M), where the implementation supports the variant numerically. All 13 families are screened at the primary setting of each equation in Section~\ref{sec:family-screen}, and the three configurations examined in detail in Section~\ref{sec:results} are members of this same set.

\section{Experimental setup}
\label{sec:setup}
\paragraph{Problems and constraint transforms.}
We study three model problems, pairing each equation with a hard constraint transform for its solution.

\emph{(i) Harmonic Oscillator:}
The HO problem is the undamped nonlinear oscillator
\begin{equation}
u''(t)+\sin(u(t))=0,\qquad u(0)=0,\quad u'(0)=1,
\end{equation}
on $[0,L]$, with $L\in\{\pi,2\pi,4\pi\}$ and a numerical reference trajectory. Lengthening $L$ adds oscillations without changing the operator. We impose the initial conditions exactly using
\begin{equation}
u(t)=t+t^2N_\theta(t).
\end{equation}

\emph{(ii) Heat Equation:}
The heat problem is
\begin{equation}
u_t-u_{xx}=0,\qquad u^*(x,t)=\sin(x)e^{-t},
\end{equation}
on $x\in[0,L]$, $t\in[0,1]$, with $L\in\{\pi,2\pi,4\pi,8\pi\}$. At $L=8\pi$, the target contains four spatial periods. We enforce the initial and boundary conditions using
\begin{equation}
u(x,t)=\sin(x)+t\,x(L-x)N_\theta(x,t).
\end{equation}

\emph{(iii) 4D manufactured problem:}
The manufactured fourth-order problem is
\begin{equation}
\frac{\partial^4u}{\partial x_1\partial x_2\partial x_3\partial x_4}-10=0,
\qquad
u^*(x)=10\prod_{k=1}^{4}x_k,
\end{equation}
on $[0,L]^4$, with $L\in\{\pi,2\pi,4\pi\}$ \citep{han2018solving,kharazmi2021hp}. Its target network output is the constant $N_\theta^*=10$. We impose the boundary conditions using
\begin{equation}
u(x)=\left(\prod_{k=1}^{4}x_k\right)N_\theta(x).
\end{equation}

In all three cases, the constraints are exact by construction, so the objective contains only the residual term and requires no loss weighting.

\paragraph{Baseline.} Throughout the matched comparisons, the baseline is the standard ungated fully connected PINN of \citet{raissi2019physics}. It uses the same architecture, residual, boundary-condition transform, collocation scheme, optimizer, training budget, and seeds as the corresponding proposed localized model, and differs from it only in Eq.~\ref{eq:localized}, which for the baseline is $g\equiv1$. Every comparison reported in Tables~\ref{tab:ho-results}--\ref{tab:family-screen} is paired in this sense, thereby isolating the effect of the first-layer localization functions. 

\paragraph{Models and setup.} HO and Heat use two hidden layers of width 64; 4D uses width 32. HO uses sine activations, which suit an oscillatory trajectory \citep{sitzmann2020implicit}, and the other two problems use tanh. The complete study contains 24{,}960 configurations and 249{,}600 seed runs across the three problems. The domain and training-epoch settings for the results reported here are summarized in Table~\ref{tab:experiment-grid}.

\begin{table}[!htbp]
\centering
\caption{Domain sizes and epoch budgets used in the localization campaign.}
\label{tab:experiment-grid}
\setlength{\tabcolsep}{4pt}
\begin{tabular}{lll}
\toprule
Problem & Domain values & Epoch budgets \\
\midrule
HO
& $t\in[0,L],\ L\in\{\pi,2\pi,3\pi,4\pi\}$
& $1$k, $3$k, $5$k, $10$k, $50$k, $100$k \\

Heat
& \shortstack[l]{$x\in[0,L],\ t\in[0,1]$\\
$L\in\{\pi,2\pi,4\pi,8\pi\}$}
& $1$k, $3$k, $5$k, $10$k, $50$k \\

4D
& $x\in[0,L]^4,\ L\in\{\pi,2\pi,4\pi\}$
& $1$k, $3$k, $5$k, $10$k \\
\bottomrule
\end{tabular}
\end{table}

\paragraph{Metrics.} We report final residual loss, residual RMSE, solution RMSE, and solution MAE. Evaluation uses the checkpoint with the lowest evaluated residual RMSE, checked at the first epoch and at regular intervals thereafter. Solution errors are measured against the numerical HO reference or the exact Heat and 4D solutions. Reported values are means over the ten seeds. For the localization family screen we use mean solution MSE, the mean across seeds of each seed's squared solution RMSE. We keep residual and solution metrics separate throughout, because the residual is what training minimizes, whereas the solution error is what the method is for, and the two can be decoupled \citep{krishnapriyan2021characterizing}.

\section{Results}
\label{sec:results}
\subsection{Harmonic oscillator (HO)}
\label{sec:ho}
\begin{figure}[!htbp]
  \centering
  \includegraphics[width=\linewidth]{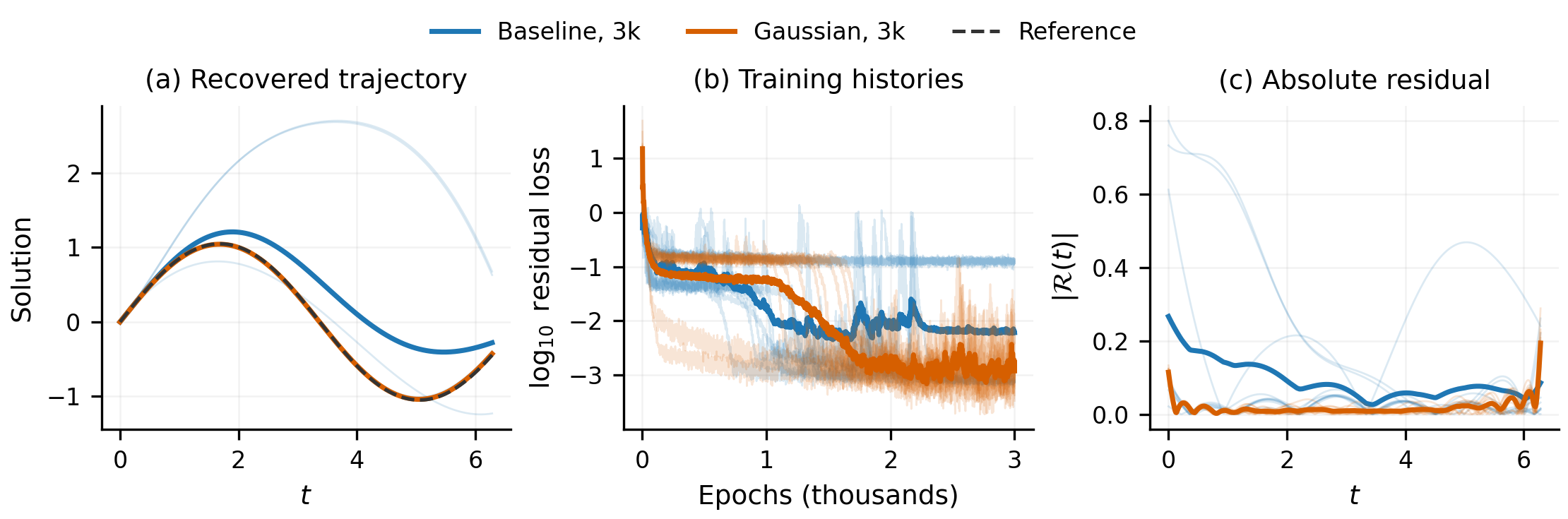}
  \caption{HO at $L=2\pi$ and 3k epochs: recovered trajectories with the numerical reference (left), residual-loss histories (center), and absolute residuals (right). Thin curves are the ten paired seeds and thick curves are their means, with loss averaged in $\log_{10}$ space.}
  \label{fig:ho-evidence}
\end{figure}

As reported in Table~\ref{tab:ho-results}, at $L=2\pi$ and 3k epochs, fixed Gaussian localization functions reduce mean solution RMSE by $98.2\%$ and MAE by $98.3\%$ relative to the baseline. Residual RMSE falls by $78.0\%$ and the final residual loss by $96.3\%$. Figure~\ref{fig:ho-evidence} shows the same improvement visually: the localized trajectory stays close to the numerical reference across the interval, while the baseline deviates more strongly; the training-history and absolute-residual panels show the corresponding lower residuals. All ten paired seeds improve in solution RMSE and MAE. With ten independent pairs, a unanimous outcome corresponds to an exact one-sided sign-test $p=2^{-10}\approx0.001$. We use this test rather than an interval on the mean because the seed-to-seed spread of the baseline is large, as Section~\ref{sec:family-screen} quantifies.

At 1k epochs the ordering reverses: the localized model is worse on all four metrics, with solution RMSE $3.2\times$ the baseline's. We report the short budget because this sensitivity is a property of the method rather than an artifact to be tuned away.

\begin{table}[!htbp]
\centering
\caption{HO at $L=2\pi$ with $64\times64$ networks and fixed Gaussian localization functions. Means over ten paired seeds. All metric entries are shown in scientific notation. Bold marks the localized solution RMSE and MAE at the larger budget.}
\label{tab:ho-results}
\setlength{\tabcolsep}{4pt}
\begin{tabular}{llrrrr}
\toprule
\shortstack[l]{Localization\\ Function} & Epochs & \hd{Final Residual\\Loss} & \hd{Residual\\RMSE} & \hd{Solution\\RMSE} & \hd{Solution\\MAE} \\
\midrule
Baseline & 3k & $5.010\times10^{-2}$ & $1.1995\times10^{-1}$ & $4.8369\times10^{-1}$ & $4.0788\times10^{-1}$ \\
Gaussian & 3k & $1.85\times10^{-3}$ & $2.639\times10^{-2}$ & \best{$8.83\times10^{-3}$} & \best{$7.06\times10^{-3}$} \\
Baseline & 1k & $6.230\times10^{-2}$ & $1.7438\times10^{-1}$ & $6.7583\times10^{-1}$ & $5.5689\times10^{-1}$ \\
Gaussian & 1k & $1.1420\times10^{-1}$ & $3.0155\times10^{-1}$ & $2.14721\times10^{0}$ & $1.79705\times10^{0}$ \\
\bottomrule
\end{tabular}
\end{table}

\FloatBarrier

\subsection{Heat equation on a long spatial domain}
\begin{figure}[!htbp]
  \centering
  \includegraphics[width=\linewidth]{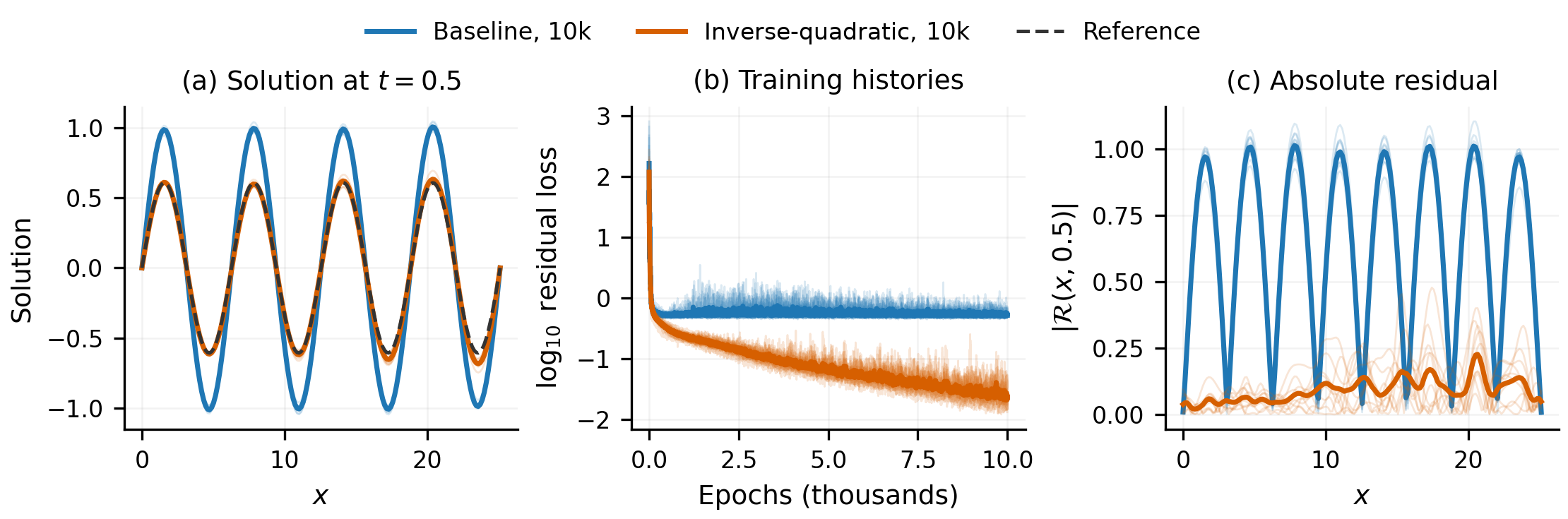}
  \caption{Heat equation at $L=8\pi$ and 10k epochs for both models: solution at $t=0.5$ (left), residual-loss histories (center), and absolute residuals at $t=0.5$ (right). Thin curves are the ten paired seeds and thick curves are their means, with loss averaged in $\log_{10}$ space.}
  \label{fig:heat-evidence}
\end{figure}

Table~\ref{tab:heat-results} reports that at $L=8\pi$ and 10k epochs, inverse-quadratic localization functions with learnable centers and widths reduce solution RMSE by $89.4\%$ and MAE by $90.3\%$ relative to the baseline. Residual RMSE falls by $78.7\%$ and the final residual loss by $95.2\%$. Figure~\ref{fig:heat-evidence} shows the same improvement visually: at $t=0.5$, the localized solution follows the reference more closely than the baseline, while the training-history and absolute-residual panels show lower residuals for the localized model. All four reported metrics improve in all ten paired seeds. The two budgets are informative together: extending the baseline from 3k to 10k epochs yields only a $0.4\%$ improvement, whereas the localized model improves by $69.7\%$; the localized model at 3k is already better than the baseline at 10k. Thus, the baseline's poorer solution accuracy is not explained simply by insufficient training.

\begin{table}[!htbp]
\centering
\caption{Heat at $L=8\pi$ with $64\times64$ networks and inverse-quadratic localization functions with learnable centers and widths. Means over ten paired seeds. All metric entries are shown in scientific notation. Bold marks the localized solution RMSE and MAE at the larger budget.}
\label{tab:heat-results}
\setlength{\tabcolsep}{4pt}
\begin{tabular}{llrrrr}
\toprule
\shortstack[l]{Localization\\Function} & Epochs & \hd{Final Residual\\Loss} & \hd{Residual\\RMSE} & \hd{Solution\\RMSE} & \hd{Solution\\MAE} \\
\midrule
Baseline & 3k & $6.433\times10^{-1}$ & $7.073\times10^{-1}$ & $2.909\times10^{-1}$ & $2.328\times10^{-1}$ \\
Inverse-quadratic & 3k & $1.208\times10^{-1}$ & $3.353\times10^{-1}$ & $1.011\times10^{-1}$ & $7.17\times10^{-2}$ \\
Baseline & 10k & $5.212\times10^{-1}$ & $7.047\times10^{-1}$ & $2.896\times10^{-1}$ & $2.319\times10^{-1}$ \\
Inverse-quadratic & 10k & $2.51\times10^{-2}$ & $1.499\times10^{-1}$ & \best{$3.06\times10^{-2}$} & \best{$2.24\times10^{-2}$} \\
\bottomrule
\end{tabular}
\end{table}

\FloatBarrier

\subsection{Four-dimensional fourth-order problem}
\label{sec:fourd}
Table~\ref{tab:four-d-results} shows that at $L=4\pi$ and 10k epochs, fixed bump localization functions reduce solution RMSE by $98.7\%$ and MAE by $99.1\%$ relative to the baseline. Residual RMSE falls by $95.3\%$, and these three evaluation metrics improve in all ten paired seeds.

\begin{figure}[!htbp]
  \centering
  \includegraphics[width=\linewidth]{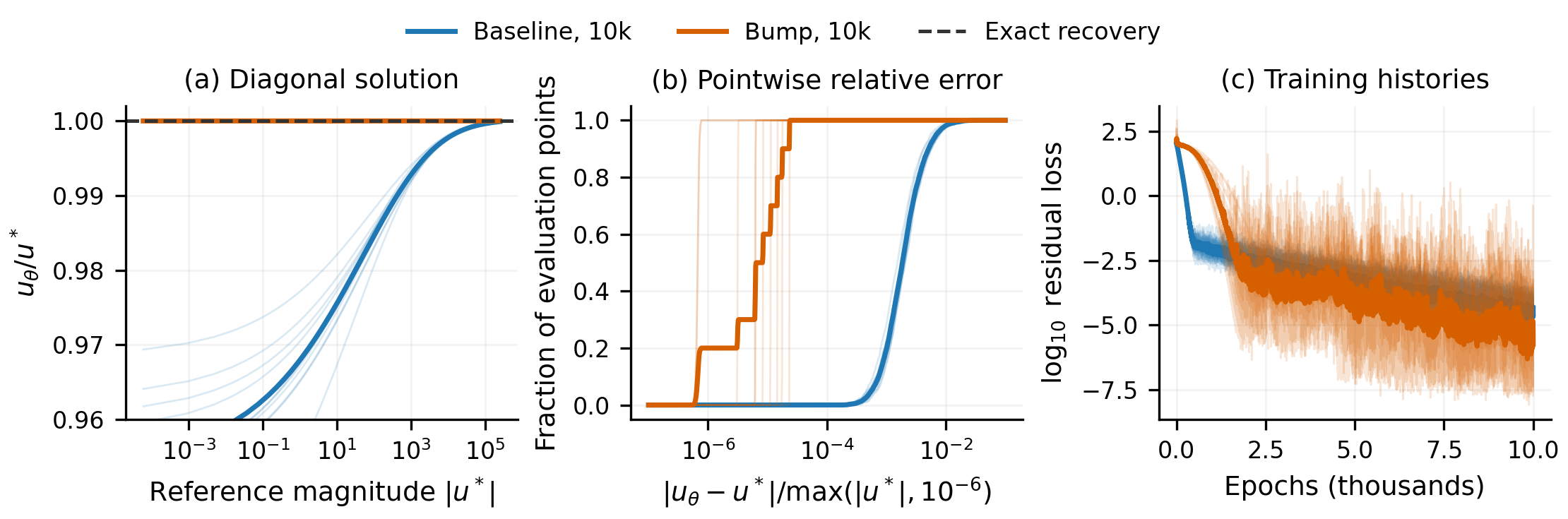}
  \caption{4D problem at $L=4\pi$ and 10k epochs for both models: normalized solution on the diagonal slice (left), empirical distribution of relative error (center), and residual-loss histories (right). Thin curves are the ten paired seeds and thick curves are their means, with loss averaged in $\log_{10}$ space.}
  \label{fig:four-d-evidence}
\end{figure}

This comparison also separates residual from solution accuracy. The final residual loss differs by only $11.6\%$, whereas the mean solution RMSE differs by a factor of $78$. Figure~\ref{fig:four-d-evidence} shows the same contrast visually: on the diagonal slice, the localized solution stays close to exact recovery, while the pointwise relative-error panel shows a larger fraction of localized evaluations at smaller errors. The training histories are much closer together than the solution-error results, illustrating why residual loss alone is not sufficient here. The baseline therefore achieves a small residual while still producing a substantially less accurate solution relative to the target scale $u^*=10\prod_k x_k$.

\begin{table}[!htbp]
\centering
\caption{4D at $L=4\pi$ with $32\times32$ networks and fixed bump localization functions. Means over ten paired seeds. All metric entries are shown in scientific notation; the exponents are given inside the cells. Bold marks the localized solution RMSE and MAE at the larger budget.}
\label{tab:four-d-results}
\setlength{\tabcolsep}{4pt}
\begin{tabular}{llrrrr}
\toprule
\shortstack[l]{Localization\\ Function} & Epochs & \hd{Final Residual\\Loss} & \hd{Residual\\RMSE} & \hd{Solution\\RMSE} & \hd{Solution\\MAE} \\
\midrule
Baseline & 3k & $2.6200\times10^{-3}$ & $5.3494\times10^{-2}$ & $1.420850\times10^{2}$ & $1.111457\times10^{2}$ \\
Bump & 3k & $2.5200\times10^{-3}$ & $1.114\times10^{-3}$ & $9.221\times10^{-1}$ & $5.287\times10^{-1}$ \\
Baseline & 10k & $5.34\times10^{-5}$ & $7.320\times10^{-3}$ & $1.75947\times10^{1}$ & $1.40520\times10^{1}$ \\
Bump & 10k & $4.72\times10^{-5}$ & $3.44\times10^{-4}$ & \best{$2.260\times10^{-1}$} & \best{$1.296\times10^{-1}$} \\
\bottomrule
\end{tabular}
\end{table}

\FloatBarrier

\section{Localization-function family screen}
\label{sec:family-screen}
Table~\ref{tab:family-screen} reports mean solution MSE for all 13 families of localization functions at the primary setting of each equation. Three properties of the table govern how it should be read. First, each family entry is the minimum over the parameterizations that were finite, whereas the baseline row involves no such minimum, so the family entries are biased low relative to the baseline. Second, mean MSE averages squared errors, so a single diverged seed can dominate a cell. Third, the three columns have different solution scales and are not comparable across columns. We therefore treat only order-of-magnitude differences within a column as evidence, and rely on the paired-seed win counts of Section~\ref{sec:results} for the detailed comparisons.

This ablation study highlights three main findings. First, localization benefits depend on the function family and the equation, with several families providing substantial gains over the baseline. On HO only Gaussian and inverse-quadratic beat the baseline; of the remaining $11$, ten are $9$ to $23$ times worse and Cauchy is $1.4$ times worse. On Heat $11$ of $13$ families are nominally better, but eight of those lie within $30\%$ of the baseline, which the selection bias above is enough to explain, leaving Cauchy, inverse-quadratic, and triangular as the only substantial gains. On 4D problem, four families are non-finite for every parameterization tested, five land between $8\times10^{3}$ and $6\times10^{4}$ times the baseline error, and four beat it. The variation across families is more pronounced for a higher-order problem, making the selection of a localization function suited to the equation important.

Second, inverse-quadratic localization is the most consistent family across the three equations. It beats the baseline on all three equations and gives the lowest entry for HO and Heat. On 4D, bump performs best, while the inverse-quadratic entry is $17.6$ times the bump entry and still $315$ times below the baseline. These results make inverse-quadratic localization functions with learnable centers and widths a strong default candidate for future experiments.

Third, the detailed comparisons in Section~\ref{sec:results} provide representative examples from the broader family screen rather than selecting its single best entry. The screen also contains stronger configurations: its lowest HO entry is inverse-quadratic at $1.963\times10^{-5}$, which is $4.3$ times below the fixed Gaussian configuration featured in Section~\ref{sec:ho}, while its lowest 4D bump entry uses learnable widths rather than the fixed parameters of Section~\ref{sec:fourd}. The featured settings were chosen because they produce large and unanimous improvements across seeds, while the broader screen identifies further configurations for future study.

The non-finite entries are recorded rather than replaced by a large finite value. For the Gaussian, Ricker, Gabor, and Morlet families on the 4D problem, the saved records hold non-finite (NaN) loss and metric values already at the first recorded checkpoint, so no finite training state was captured for those runs; this is a missing measurement, not a small or zero loss. All four families are finite on HO and Heat and fail only here, where the residual differentiates the localization functions four times. The records establish the numerical failure, but not its precise mechanism. Repeated higher-order differentiation of a localized activation can create underflow, overflow, vanishing gradients, or other instabilities that prevent a finite checkpoint from being recorded. No equation-specific stabilization was added for the fourth-order residual, because the study was intended to keep one generalized implementation across all three equations.

\begin{table}[!htbp]
\centering
\caption{Screen of localization-function families at the primary settings: HO at $2\pi$/3k, Heat at $8\pi$/10k, and 4D at $4\pi$/10k. Entries are mean solution MSE over ten seeds; lower is better. F, S, and M denote fixed, learnable-width, and learnable-center/width parameterizations; each cell gives the lowest finite parameterization for that family, with the winning variant in parentheses. n/a means that no tested parameterization produced a finite solution metric for that problem. Bold marks the lowest valid entry in each problem column. Columns have different solution scales and should be compared only within a column.}
\label{tab:family-screen}
\begin{tabular}{l|c|c|c}
\toprule
Localization Function & HO & Heat & 4D \\
\midrule
Baseline & $9.946\times10^{-1}$ & $8.386\times10^{-2}$ & $3.126\times10^{2}$ \\
\midrule
Gaussian & $8.402\times10^{-5}$ (F) & $7.956\times10^{-2}$ (F) & n/a \\
Activated & $1.788\times10^{1}$ (S) & $7.502\times10^{-2}$ (S) & $1.777\times10^{7}$ (F) \\
Super-Gaussian & $1.541\times10^{1}$ (S) & $7.470\times10^{-2}$ (M) & $1.621\times10^{7}$ (S) \\
Ricker & $1.148\times10^{1}$ (F) & $8.339\times10^{-2}$ (F) & n/a \\
Boxcar & $9.140$ (S) & $7.237\times10^{-2}$ (S) & $1.346\times10^{7}$ (S) \\
Laplace & $1.509\times10^{1}$ (M) & $7.895\times10^{-2}$ (M) & $7.892\times10^{1}$ (F) \\
Cauchy & $1.435$ (M) & $1.640\times10^{-3}$ (M) & $3.613$ (F) \\
Raised-cosine & $9.169$ (M) & $7.162\times10^{-2}$ (S) & $7.233\times10^{6}$ (M) \\
Bump & $1.727\times10^{1}$ (F) & $6.557\times10^{-2}$ (S) & \best{$5.650\times10^{-2}$ (S)} \\
Gabor & $1.862\times10^{1}$ (F) & $8.491\times10^{-2}$ (F) & n/a \\
Morlet & $2.280\times10^{1}$ (F) & $8.410\times10^{-2}$ (F) & n/a \\
Inverse-quadratic & \best{$1.963\times10^{-5}$ (M)} & \best{$9.780\times10^{-4}$ (M)} & $9.924\times10^{-1}$ (M) \\
Triangular & $1.391\times10^{1}$ (M) & $1.176\times10^{-2}$ (S) & $2.598\times10^{6}$ (S) \\
\bottomrule
\end{tabular}
\end{table}
\FloatBarrier

Finally, the results together recover the across-seed spread that neither shows on its own, because mean MSE and mean RMSE are the first two moments of the same ten values. For the HO baseline they imply a standard deviation of $8.7\times10^{-1}$ around a mean solution RMSE of $4.8369\times10^{-1}$, against $2.5\times10^{-3}$ around $8.83\times10^{-3}$ for the fixed Gaussian model. The baseline's seed-to-seed variation is thus larger than its own mean error, which is the reason we report unanimity across pairs instead of an interval on the mean.

\section{Conclusion and future work}

Across the three tested problems, first-layer localization improves recovered solution accuracy while leaving the PINN, residual, optimizer, and boundary transform unchanged, and the improvement is reproduced in every paired seed of the detailed comparisons. The 13-family screen shows that localization is a flexible design choice whose benefit depends on matching the family to the equation and derivative order: inverse-quadratic localization with learnable centers and widths is the most consistent configuration across the three equations, while the broader screen also identifies additional family--problem combinations worth exploring. Because the study uses matched comparisons and controlled analytic benchmarks, it isolates the optimization effect of localization and provides a clear foundation for testing the approach on broader equations and more realistic scientific settings.

Future work should extend the study to a broader range of equations and more realistic scientific applications, moving beyond the controlled benchmark problems considered here. This will test whether the benefits of first-layer localization persist for nonseparable targets, coupled systems, and real-world problem settings.

\subsection*{Reproducibility statement}
Sections~\ref{sec:method} and~\ref{sec:setup} define the architecture, the localization functions with their families and parameterizations, the problems, constraint transforms, widths, activations, budgets, seeds, and checkpoint rule used for every reported number. Baseline and localized models share all settings except the bank of localization functions in Eq.~\ref{eq:localized}. The experiments were run on a 16-inch Apple M3 Max MacBook Pro with 36\,GB RAM, CPU-only multiprocessing. The source code will be available in the public Git repository. 

\subsection*{AI use statement}
Generative AI tools were used for code-assisted editing, formatting, figure preparation, and language revision. The authors reviewed and verified the AI-assisted text, equations, citations, reported results, figures, and code. Generative AI was not used to generate experimental data, to select which results to report, or to replace author verification of the reported results. The authors take responsibility for the final content.

\bibliography{references}
\bibliographystyle{iclr2027_conference}

\end{document}